\documentclass[letterpaper, 10pt, journal]{ieeeconf}

\IEEEoverridecommandlockouts                             
\usepackage{times}

\usepackage{multicol}
\usepackage[bookmarks=true]{hyperref}
\usepackage{graphicx}
\usepackage{support-caption}
\usepackage{subcaption}
\usepackage{tabularx}
\usepackage{amsmath}
\usepackage{amssymb}
\usepackage{bm}
\usepackage[normalem]{ulem}
\usepackage{makecell}
\usepackage{hhline}
\usepackage{siunitx}
\usepackage{blkarray}
\usepackage{boldline}
\usepackage{siunitx}
\usepackage{float}

\usepackage[
backend=bibtex, 
bibencoding=utf8,
style=ieee, 
sorting=none, 
natbib=true, 
doi=false, 
isbn=false, 
url=false, 
eprint=false, 
maxcitenames=1, 
mincitenames=1,
dashed=false
]{biblatex}

\usepackage{xspace}

\usepackage[printonlyused]{acronym}
\acrodef{ICP}{iterative closest point}
\acrodef{NDT}{normal distribution transformation}
\acrodef{SA}{Set Abstraction}
\acrodef{FP}{Feature Propagation}
\acrodef{GNSS}{Global Navigation Satellite System}
\acrodef{HMI}{Hazardously Misleading Information}

\usepackage{lipsum}
\usepackage[dvipsnames]{xcolor}

\AtBeginDocument{}%
\AtBeginDocument{}%

\title{\LARGE \bf
Prepared for the Worst: Resilience Analysis of the ICP Algorithm via Learning-Based Worst-Case Adversarial Attacks
}

\author{Ziyu Zhang$^{1}$, Johann Laconte$^{1}$, Daniil Lisus$^{1}$, and Timothy D. Barfoot$^{1}$ 
\thanks{$^{1}$ University of Toronto Institute for Aerospace Studies (UTIAS), 4925 Dufferin St, Ontario, Canada.
\tt\small char.zhang@robotics.utias.utoronto.ca%
}
}

\begin{document}

\newpage
%
%
%
%
%
%
%
\def \myJournal {IEEE International Conference on Robotics and Automation (ICRA)}
\def \myDoi {10.1109/ICRA55743.2025.11128007}
\def \myPaperSiteName {IEEE Xplore}
\def \myPaperSiteLink {https://ieeexplore.ieee.org/document/11128007}
\def \myYear {2025}

\def \myPaperCitation{Z. Zhang, J. Laconte, D. Lisus and T. D. Barfoot, "Prepared for the Worst: Resilience Analysis of the ICP Algorithm via Learning-Based Worst-Case Adversarial Attacks," 2025 IEEE International Conference on Robotics and Automation (ICRA), Atlanta, GA, USA, 2025, pp. 15174-15180.}


\begin{figure*}[t]

\thispagestyle{empty}
\begin{center}
\begin{minipage}{6in}
\centering
This paper has been accepted for publication in \emph{\myJournal}. 
\vspace{1em}

This is the author's version of an article that has, or will be, published in this journal or conference. Changes were, or will be, made to this version by the publisher prior to publication.
\vspace{2em}

\begin{tabular}{rl}
DOI: & \myDoi\\
\myPaperSiteName: & \texttt{\myPaperSiteLink}
\end{tabular}

\vspace{2em}
Please cite this paper as:

\myPaperCitation

\vspace{15cm}
\copyright \myYear \hspace{4pt}IEEE. Personal use of this material is permitted. Permission from IEEE must be obtained for all other uses, in any current or future media, including reprinting/republishing this material for advertising or promotional purposes, creating new collective works, for resale or redistribution to servers or lists, or reuse of any copyrighted component of this work in other works.

\end{minipage}
\end{center}
\end{figure*}
\newpage
\clearpage
\pagenumbering{arabic} 

\maketitle

\markboth{IEEE International Conference on Robotics and Automation (ICRA). Preprint. Accepted Jan, 2025}{Zhang \MakeLowercase{\textit{et al.}}: Prepared for the Worst: Resilience Analysis of the ICP Algorithm via Learning-Based Worst-Case Adversarial Attacks}

\thispagestyle{headings}
\pagestyle{empty}

\begin{abstract}

This paper presents a novel method for assessing the resilience of the \ac{ICP} algorithm via learning-based, worst-case attacks on lidar point clouds.
For safety-critical applications such as autonomous navigation, ensuring the resilience of algorithms before deployments is crucial. 
The \ac{ICP} algorithm is the standard for lidar-based localization, but its accuracy can be greatly affected by corrupted measurements from various sources, including occlusions, adverse weather, or mechanical sensor issues.
Unfortunately, the complex and iterative nature of \ac{ICP} makes assessing its resilience to corruption challenging. While there have been efforts to create challenging datasets and develop simulations to evaluate the resilience of \ac{ICP}, our method focuses on finding the maximum possible \ac{ICP} error that can arise from corrupted measurements at a location.
We demonstrate that our perturbation-based adversarial attacks can be used pre-deployment to identify locations on a map where \ac{ICP} is particularly vulnerable to corruptions in the measurements. With such information, autonomous robots can take safer paths when deployed, to mitigate against their measurements being corrupted.
The proposed attack outperforms baselines more than 88\% of the time across a wide range of scenarios.
\end{abstract}

\section{Introduction}
    \label{sec:intro}
\begin{figure}
    \centering
    \begin{subfigure}[b]{0.49\textwidth}
        \centering
        \includegraphics[width=0.97\textwidth]{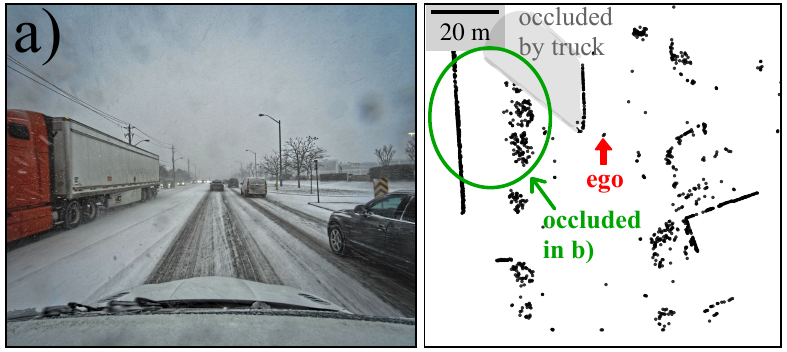}
        \label{fig:sub1}
    \end{subfigure}
    \begin{subfigure}[b]{0.49\textwidth}
        \centering
        \includegraphics[width=0.97\textwidth]{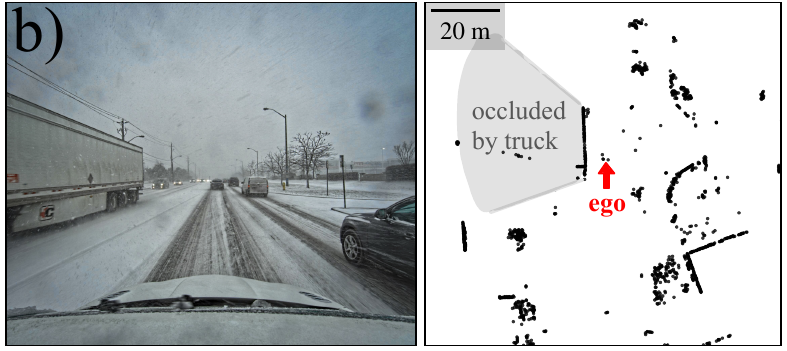}
        \label{fig:sub2}
    \end{subfigure}
    \caption{{Illustrative scenarios from the Boreas dataset \cite{burnett_ijrr23} where the ego vehicle is occluded by a truck.} Camera images (left) and lidar scans (right) depict two sequential scenarios, with scenario a) preceding b) by \SI{0.3}{\s}. The ego vehicle’s pose (red arrow) and occluded region (gray sector) are annotated. When the truck occludes the green-circled landmarks in b), the \ac{ICP} algorithm may mistake the truck for the circled landmarks, resulting in significant errors in the pose estimate. }
    \label{fig:overall}
\end{figure}

    The iterative closest point (ICP) algorithm has become a fundamental localization algorithm in mobile robotics \cite{besl_and_mckay} \cite{review_ICP}.
\ac{ICP} computes a robot’s current pose by determining the transformation that optimally aligns the scan point cloud (robot’s current view) with a map point cloud. 
Meanwhile, lidar sensors have emerged as the predominant choice for robot localization and mapping \cite{yin2023survey}.
However, despite its popularity, lidar-based \ac{ICP} is prone to failures when the measurements are corrupted, such as during significant occlusions or adverse weather.
Figure \ref{fig:overall} illustrates how landmarks needed for localization can be occluded in typical autonomous driving conditions.
Once the real landmarks are occluded, \ac{ICP} may attempt to match the obstacle itself to the map, resulting in unexpected pose errors.
Similar errors can arise in adverse weather conditions. 
For instance, when the wind carries snow into the proximity of landmarks, it becomes difficult for \ac{ICP} to distinguish and filter the snow from the landmarks \cite{courcelle2022importance}.
Corrupted measurements pose a significant safety threat. This paper aims to quantify how susceptible a given map is to this threat as a way to evaluate the safety of different regions of the map prior to deployment.

To evaluate the resilience of \ac{ICP} against corrupted measurements, adverse weather datasets were collected \cite{burnett_ijrr23} \cite{Pitropov_2020}. Prior works \cite{courcelle2022importance} \cite{burnett2023ready} have evaluated \ac{ICP} on these datasets.
\citet{s21155196} simulated occlusion cases to evaluate the performance of lidar-based localization algorithms.
Our paper presents a new tool for \ac{ICP} resilience analysis, which estimates the maximum pose error that may arise from corrupted measurements at a given map location. Finding the maximum possible error evaluates the safety of a map under the worst conditions, instead of some specific, hard-to-model, `realistic' conditions. This approach abstracts away all possible corruptions when evaluating a map, providing a lower bound on localization safety.

While \citet{laconte2023certifying} presented an analytic approach for this task, we propose a learning-based approach that requires fewer assumptions. 
Following \cite{laconte2023certifying}, we abstract the problem by modeling corruption as measurement perturbations.
We train an attack network to maximize the \ac{ICP} error via perturbing points in a lidar scan. Once trained, the maximum possible pose error at a location, subject to measurement perturbations, can be estimated by querying our network with a lidar scan of the location. 
Before deploying autonomous robots, previously collected scans or maps can be fed into the network. The network can help identify, offline, dangerous map locations where the worst-case pose error exceeds a tolerance, allowing these locations to be avoided during deployment.
We evaluate our approach by attacking point-to-plane \ac{ICP} on the ShapeNetCore \cite{chang2015shapenet} and Boreas \cite{burnett_ijrr23} datasets. 
Results show that our attack introduces measurement perturbations \ac{ICP} struggles to filter, resulting in large \ac{ICP} errors. Using this attack, we detect many dangerous map locations the current state-of-the-art method misses.
Our contributions are as follows.
\begin{itemize}
    \item To the best of our knowledge, we propose the first learning-based adversarial attack on lidar-based \ac{ICP}.
    \item We present a novel way to estimate the worst \ac{ICP} pose error that can arise at a location subject to a given amount of perturbation in the point cloud.
    \item We demonstrate the feasibility of using our approach to evaluate map safety and identify, pre-deployment, dangerous locations to avoid.
\end{itemize}
\section{Related Work}
    \label{sec:rw}
    \subsection{Analysis of ICP in Challenging Environments}
Geometrically under-constrained environments refer to environments with very limited or degenerate geometric constraints for registration such as tunnels and narrow corridors. Under-constrained environments have long been established as a major source of \ac{ICP} error \cite{censi2007}. Numerous degeneracy detection techniques and degeneracy-aware localization methods have been proposed \cite{tuna2023xicp}.
Scenarios such as occlusions and adverse weather, on the other hand, are challenging due to high levels of corruption in the lidar measurements. Corrupted measurements pose a significant safety threat, and they are the focus of this paper.

To make registration algorithms, including \ac{ICP}, robust to noisy measurements, numerous outlier filters have been proposed \cite{babin}.
Another line of research suggests to address this problem via de-noising the measurements. \citet{charron} presented a method to remove noise in 3D lidar point clouds caused by snow.
This paper approaches the issue from a different angle, aiming to quantify and understand the impact of corrupted measurements on lidar-based \ac{ICP}.

\citet{burnett2023ready} tested lidar-based localization system on the all-weather Boreas dataset \cite{burnett_ijrr23}. They found that lidar-based localization is surprisingly robust to moderate precipitation. Later, \citet{courcelle2022importance} evaluated lidar-based \ac{ICP} over the Canadian Adverse Driving Conditions (CADC) dataset \cite{Pitropov_2020} and also discovered that lidar-based \ac{ICP} is robust to high levels of precipitation. However, they observed that extreme, abrupt cases such as snow gusts led to significant \ac{ICP} localization errors. Due to an insufficient amount of data on these extreme cases, they could not quantitatively evaluate the impact of these extreme events on \ac{ICP}. 
\citet{courcelle2022importance} identified view obstruction (occlusion) as another event that led to large pose errors. \citet{s21155196} presented an evaluation of lidar-based localization under occlusions via simulation. Unsurprisingly, they found that, on average, localization error increases with the percentage of the scene being occluded and significant localization errors can arise when half or more of the scene is occluded.

In our previous work, \citet{laconte2023certifying} proposed a closed-form method for estimating the maximum expected \ac{ICP} error from a sector of corrupted measurements. 
They then quantify \ac{ICP}'s resilience using the smallest sector of corruption that can cause significant errors.
This paper proposes an alternative solution to assessing \ac{ICP}'s resilience that removes three key assumptions made in \cite{laconte2023certifying}: 
While \cite{laconte2023certifying} only targets single-iteration \ac{ICP}, this paper attacks the much more commonly used multi-iteration \ac{ICP}. Second, our data association is re-calculated at every \ac{ICP} iteration, whereas \cite{laconte2023certifying} assumes a known data association. Third, while \cite{laconte2023certifying} does not consider robust filters, our approach can attack \ac{ICP} with robust filters. 
A direct comparison can be found in Section \ref{section:sota-comparison}.
Similar to \cite{laconte2023certifying}, our approach can pinpoint dangerous locations where significant localization errors can occur if lidar measurements are corrupted. Unlike degeneracy detection methods \cite{degeneracy_detection_nubert}, which predict the localizability of a location under normal conditions, our method and \cite{laconte2023certifying} account for potential measurement corruptions. A location can have enough geometric constraints yet still experience significant pose errors when the measurements are corrupted. However, the assumptions made in \cite{laconte2023certifying} cause it to overlook many dangerous map locations that our method finds.

\subsection{Adversarial Attacks Against Autonomous Driving Systems}

Adversarial attacks that deliberately craft examples to undermine the target algorithm’s performance are very suitable for our purpose of worst-case analysis. Many point cloud adversarial attacks have been proposed against the autonomy stack. \citet{yang2021adversarial} introduced novel frameworks for attacking algorithms processing 3D point clouds. These attacks work via point addition, removal, and perturbation. \citet{zhou_lg-gan_2020} proposed the first generative attack against point cloud classification algorithms via point perturbation. They showed that generative approaches are much faster than gradient-based approaches while upholding good attack performance. For this reason, and the resemblance of our task to theirs, our architecture draws inspiration from theirs. 

\citet{zhang2022adversarial} proposed an attack that perturbs vehicle trajectories to maximize the errors of trajectory prediction algorithms. They successfully increased the prediction errors by more than 150\% and showed that worst-case predictions have critical safety concerns. Their work supports our proposal of using adversarial attacks for worst-case performance analysis. For localization algorithms, prior work explored adversarial attacks on visual SLAM \cite{wang_i_2021} \cite{perceptual_aliasing}. On lidar-based SLAM, \citet{xu_sok_2023} demonstrated the feasibility of introducing false loop-closure detection by increasing the similarity between two distinct locations using physical objects. Our work differs from these works in that we are not merely introducing errors in the target algorithm but rather maximizing its error. Moreover, their target algorithms and attack mechanisms significantly differ from ours.

\citet{yoshida_adversarial_2022} proposed a method for misleading lidar-based \ac{ICP} to a specific wrong pose via adversarial point perturbation.
While both \cite{yoshida_adversarial_2022} and our work are adversarial attacks on \ac{ICP}, the attack objectives differ. \citet{yoshida_adversarial_2022} aim to corrupt the scan to mislead \ac{ICP} to output a given incorrect pose that the adversary desires. These incorrect poses are neither designed nor guaranteed to maximize \ac{ICP} errors. Our attack learns to corrupt the scan to maximize \ac{ICP} errors. Therefore, unlike \cite{yoshida_adversarial_2022}, ours can estimate the maximum pose error that may result from a given extent of perturbations in the scan.
\section{Theory}
    \label{sec:theory}

\begin{figure*}[ht]
    \centering
    \vskip 4pt
    \includegraphics[width=0.92\linewidth]{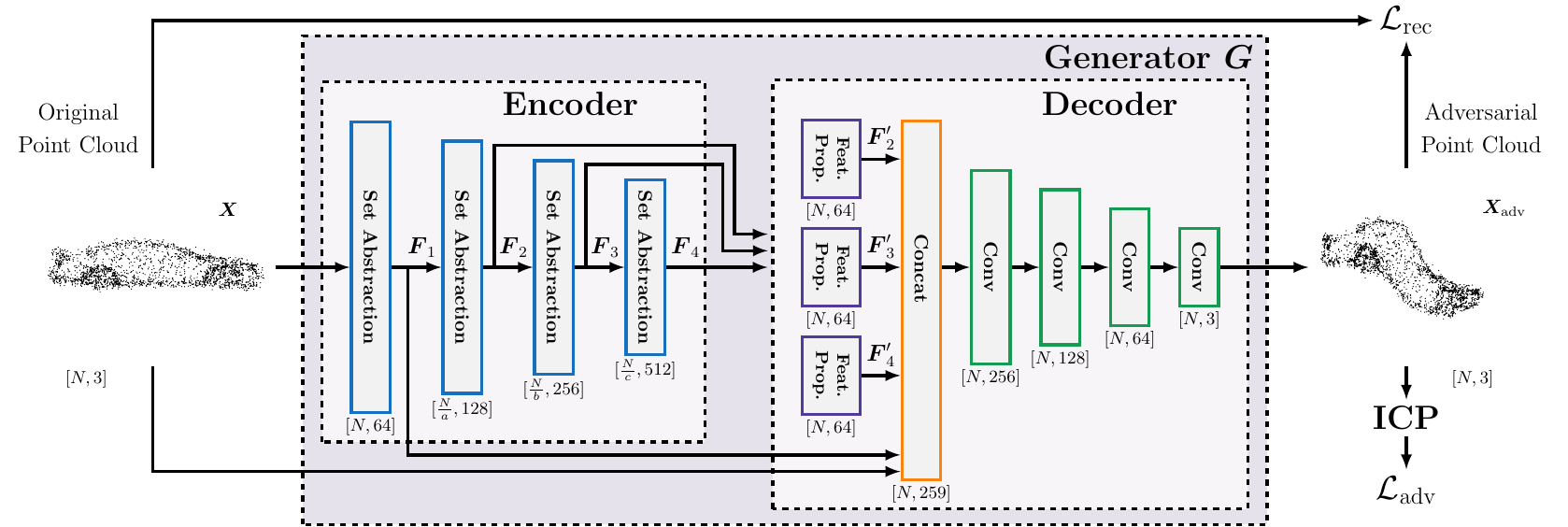}
    \caption{{Overview of the attack pipeline.} Given an input point cloud $\boldsymbol{X}$, the encoder extracts hierarchical features from it using set-abstraction modules \cite{qi2017pointnet}. The decoder interpolates the extracted features and uses them along with $\boldsymbol{X}$ to produce an adversarial point cloud $\boldsymbol{X}_{\text{adv}}$ to attack \ac{ICP}. The encoder and decoder together form the generator $\boldsymbol{G}$.}
    \label{fig:architecture}
\end{figure*}

This section details the attack target, model architecture, and loss functions used for training.
Our attack pipeline, which is the same during training and testing, is visualized in Figure \ref{fig:architecture}. We propose a generative network that learns how to perturb a point cloud to maximize the pose error of \ac{ICP} while keeping the perturbations within a specified bound.

\subsection{Attack Target: ICP}
\label{attack target}
Our model can attack any differentiable lidar-based \ac{ICP} algorithm. 
We train using dICP \cite{lisus2023pointing}, a differentiable \ac{ICP} library.
We attack the single-frame \ac{ICP} algorithm, rather than a full localization pipeline (e.g., with odometry), which we leave for future work.

Given a reference point cloud (also known as the map) \mbox{$\boldsymbol{Q} \in \mathbb{R}^{M\times 3} $} and a measured point cloud (also known as the scan) \mbox{$\boldsymbol{P} \in \mathbb{R}^{N\times 3}$}, \ac{ICP} estimates a transform from the scan to the map \mbox{${\boldsymbol{{\hat{T}}_{QP}}} \in SE(3)$}. \mbox{$M, N \in \mathbb{N}$} denote the number of points in the map and scan, respectively. The \ac{ICP} pose error vector \mbox{$\boldsymbol{\xi} \in \mathbb{R}^6$} can be calculated using the ground truth transform \mbox{$\boldsymbol{T_{QP}} \in SE(3)$} via
\begin{equation}
	\boldsymbol{\xi} = \begin{bmatrix}
	    \boldsymbol{\rho} \\
        \boldsymbol{\phi}
	\end{bmatrix} =  \log\left(\boldsymbol{{\hat{T}}_{QP}} \boldsymbol{{T_{QP}}}^{-1}\right)^\vee,
\end{equation}
where \mbox{$\boldsymbol{\rho} \in \mathbb{R}^{3}$} is the translation component of the error and \mbox{$\boldsymbol{\phi} \in \mathbb{R}^{3}$} is the rotation component of the error. Here, \mbox{$\log(\cdot)$} maps an \mbox{$SE(3)$} element to its lie algebra \mbox{$\mathfrak{se}(3)$} and \mbox{$(\cdot)^\vee$} maps an \mbox{$\mathfrak{se}(3)$} element to \mbox{$\mathbb{R}^{6}$} \cite{barfoot_state_estimation}.
For brevity, we only attack point-to-plane \ac{ICP} in this paper. However, our method could similarly be applied to other algorithms such as point-to-point \ac{ICP} and NDT \cite{NDT}.

\subsection{Attack model}
\label{attack model}
Our model is a generator $\boldsymbol{G}$ consisting of an encoder and a decoder. The encoder is based on PointNet++ \cite{qi2017pointnet} \cite{Pytorch_Pointnet_Pointnet2} and learns to extract hierarchical features from an input point cloud \mbox{$\boldsymbol{X} \in \mathbb{R}^{N \times 3}$}. Using the extracted features, the decoder learns how to perturb $\boldsymbol{X}$ to get \mbox{$\boldsymbol{X}_{\text{adv}} \in \mathbb{R}^{N \times 3}$}. 
This architecture is inspired by LG-GAN \cite{zhou_lg-gan_2020}.

\subsubsection{Encoder}
To extract hierarchical features, the encoder is designed with four cascaded PointNet++ \cite{qi2017pointnet} set-abstraction modules, as shown in Figure \ref{fig:architecture}. A set-abstraction module takes a set of points and samples a smaller set using farthest-point sampling. Every sampled point captures local patterns by learning to aggregate features of its neighboring points in the original set. This subset of points along with newly learned features are fed into the next set-abstraction module. By stacking set-abstraction modules, a hierarchy of features of various scales can be extracted. Precisely, the encoder extracts features of four scales: \mbox{$\boldsymbol{F}_{1} \in \mathbb{R}^{N \times 64}$}, \mbox{$\boldsymbol{F}_{2} \in \mathbb{R}^{\frac{N}{a} \times 128}$}, \mbox{$\boldsymbol{F}_{3} \in \mathbb{R}^{\frac{N}{b} \times 256}$}, and \mbox{$\boldsymbol{F}_{4} \in \mathbb{R}^{\frac{N}{c} \times 512}$} where \mbox{$a, b, c \in \mathbb{R}$ are manually selected hyperparameters}.

\subsubsection{Decoder} 
As higher-level features are sparser, $\boldsymbol{F}_{2}$, $\boldsymbol{F}_{3}$, and $\boldsymbol{F}_{4}$ are interpolated. The procedure for interpolation is as follows. For every point in $\boldsymbol{X}$, higher-level features of $k$-nearest neighbors are weighted inverse-proportionally to their distances and summed. These weighted sums are passed through neural network layers, which turn the features into more compact sizes of \mbox{$N \times 64$}. Finally, interpolated \mbox{$\boldsymbol{F}_{2}', \boldsymbol{F}_{3}', \boldsymbol{F}_{4}' \in \mathbb{R}^{N \times 64}$} are concatenated with $\boldsymbol{F}_{1}$ and $\boldsymbol{X}$, and the result is passed through four 1D convolution layers to generate the adversarial point cloud ${\boldsymbol{X}}_{\text{adv}}$.

\subsection{Loss function}
\label{loss function}
\label{eq:total_loss}
The loss function is a weighted sum of the adversarial loss $\mathcal{L}_{\text{adv}}$ and reconstruction loss $\mathcal{L}_{\text{rec}}$. That is
\begin{equation}
\label{total loss}
	\mathcal{L} = \alpha \mathcal{L}_{\text{adv}} + \beta \mathcal{L}_{\text{rec}},
\end{equation}
where \mbox{$\alpha, \beta \in \mathbb{R} > 0$} are manually selected hyperparameters. 
The adversarial loss guides the generator $\boldsymbol{G}$ to generate an adversarial point cloud $\boldsymbol{X}_{\text{adv}}$ that is optimized to maximize the \ac{ICP} pose error. Therefore, $\mathcal{L}_{\text{adv}}$ is defined as
\begin{equation}
\label{eq:adv_loss}
    \mathcal{L}_{\text{adv}} = - \left\| 
    \begin{bmatrix} 
    w_1 \\ 
    w_2 \\ 
    w_3 
    \end{bmatrix} 
    \odot \boldsymbol{\rho}\right\|_2 - \left\|
    \begin{bmatrix} 
    w_4 \\ 
    w_5 \\ 
    w_6 
    \end{bmatrix} 
    \odot \boldsymbol{\phi} 
    \right\|_2
\end{equation}
where $\odot$ is the element-wise multiplication operator. Manually selected hyperparameters \mbox{$w_i \in \mathbb{R}$} for \mbox{$i = \text{1, ..., 6}$} specify the weights of the pose error elements.
These weights are solely dependent on the pose elements whose resiliency the user wishes to test. Section \ref{case_study} provides a case study on selecting these weights.
The \ac{ICP} error is split into translation \mbox{$\boldsymbol{\rho} \in \mathbb{R}^{3}$} and rotation \mbox{$\boldsymbol{\phi} \in \mathbb{R}^{3}$} components due to differences in scale and physical representation.

To better assess the real-world \ac{ICP} deployment risks, we introduce a loss $\mathcal{L}_{\text{rec}}$ that binds the perturbations $\boldsymbol{G}$ can introduce. \ac{ICP} algorithms often include outlier filters, rendering extreme perturbations ineffectual in maximizing pose errors. 
However, there remain other reasons to constrain the perturbations further. For example, certain real-world events may only cause perturbations up to a realistic threshold. The reconstruction loss is defined as
\begin{equation}
\label{eq:recons_loss}
\mathcal{L}_\text{rec} = \frac{1}{N}{\sum_{i=1}^{N} S(\|\boldsymbol{x}_{i} - (\boldsymbol{x_{\text{adv}}})_{i}) \|_2)^{2}},
\end{equation}
where $\boldsymbol{x}_{i}$ is a point in {the scan} $\boldsymbol{X}$ and $(\boldsymbol{x_{\text{adv}}})_{i}$ is the corresponding point in  $\boldsymbol{X}_{\text{adv}}$. $S(\cdot)$ is the SoftShrinkage function 
\begin{equation}
\label{eq:softshrink}
S(z) = \begin{cases}
        z - \lambda, & \text{if } z > \lambda \\
        z + \lambda, & \text{if } z < -\lambda\\
        0, & \text{otherwise.}
    \end{cases}
\end{equation}
The customizable parameter $\lambda \geq 0$ represents the perturbation bound. 
Once trained, the generator $\boldsymbol{G}$ can estimate the maximum pose error that can arise at a location with up to $\lambda$ units (usually meters) of perturbation in the measurements.

\section{Experiments}
    \label{sec:experiments}
    This section presents results on the ShapeNetCore \cite{chang2015shapenet} and Boreas \cite{burnett_ijrr23} datasets. Implementation details for the two datasets differ and are documented separately. 
We compare our attack with heuristic baselines to show its strong performance in maximizing \ac*{ICP} errors. Then, we compare it with the state-of-the-art method to demonstrate the benefits of using our attack to identify dangerous map locations.
\subsection{Datasets}
\label{dataset}
ShapeNetCore, a subset of ShapeNet containing 3D models of simple objects, is used to visually analyze the perturbations learned by our model. Following \cite{wang2018pixel2mesh}, we sample points and generate normals.
We use the full point clouds as maps and randomly sample 2,048 points to create scans. As ShapeNetCore objects vary significantly in size, we normalize the point clouds to fit within a unit circle centered at the origin.
We add normally distributed noise (mean 0, standard deviation 0.025) to the scans to avoid unrealistically perfect alignment with the maps. 
A small, random transformation is applied to each scan, serving as the ground truth pose. The transformation involves uniformly distributed translations from -0.08 to 0.08 (unitless due to normalization) along the $x$- and $y$-axis and a uniformly distributed rotation from -10 to 10 degrees around the $z$-axis.
The transformations are kept small such that the \ac{ICP} algorithm can use no motion (i.e., identity) for its initial guesses.

To assess our attack's effectiveness for autonomous navigation applications, we also evaluate it on the Boreas dataset \cite{burnett_ijrr23}, an autonomous driving dataset collected by manually driving a repeated route over a year. 
We use the Teach and Repeat \cite{original_teach_repeat} framework to establish localization pairs, as described in \cite{burnett2023ready}. Teach and Repeat first conducts a teach pass along a route to construct a map, to which the subsequent repeat passes along the same route can be localized. 
Once localization pairs are generated, we preprocess them to use for training our network. First, we align the live scans with the corresponding submaps using the ground truth poses provided by the Global Navigation Satellite System. Then we apply a small, random transformation to each live scan to generate the final scan we task \ac{ICP} to localize. The transformations involve random translations, uniformly distributed from -0.3 to 0.3 meters, along the $x$-, $y$-, and $z$-axis, as well as random rotations, uniformly distributed from -10 to 10 degrees, around the $x$-, $y$-, and $z$-axis.

\subsection{Baselines}
\label{baselines}
To evaluate the ability of our method to maximize \ac{ICP} errors, we compare it with two baselines.
\subsubsection{Uniform Translation Baseline}
\label{uniform_baseline}
This baseline uniformly translates the entire scan by the maximum allowed perturbation, $\lambda$, at random angles in the $x-y$ plane. The perturbations are restricted to the $x-y$ plane because we focus on the lateral and longitudinal localization errors a model can induce. These errors are more significant than vertical pose errors in autonomous navigation applications. 
\subsubsection{Normal Translation Baseline}
\label{normal_baseline}
This baseline moves points in the scan by $\lambda$ in the direction of their normal vectors projected onto the $x-y$ plane. Shifting measured points along the normal vectors of their associated map points is shown to be an effective attack on point-to-plane \ac{ICP} in \cite{laconte2023certifying}.
Directions of the normal vectors are unified such that normals of the same geometric feature point in the same general direction.

\begin{figure*}[ht]
  \vskip 5pt
  \includegraphics[width=\linewidth]{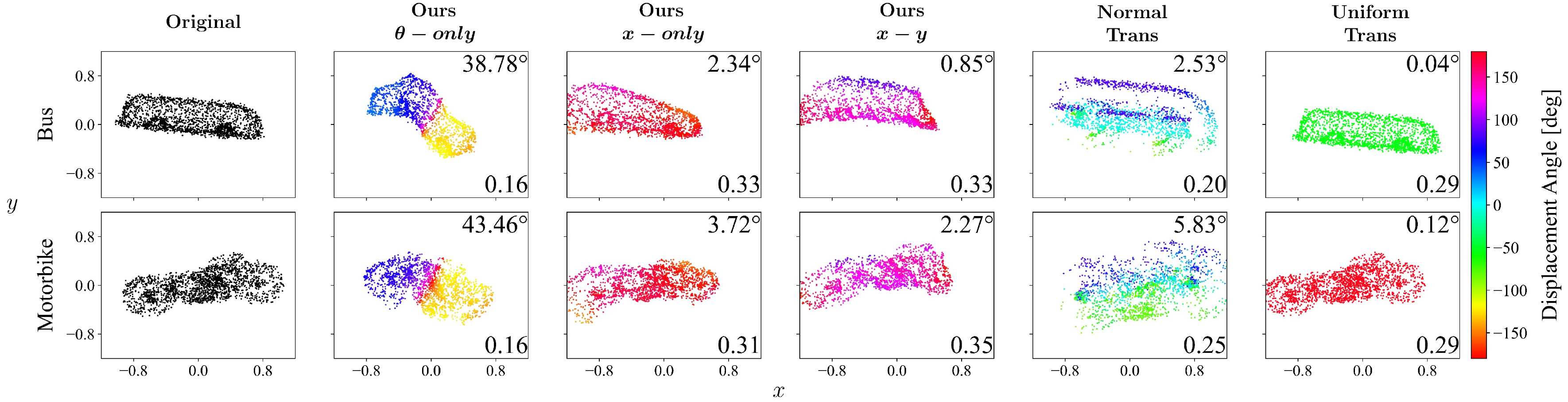}
\caption{{Original point clouds and the adversarial point clouds produced by our model and the baselines.} Columns 2 to 6 are the adversarial point clouds resulting from applying our models and the baselines to the original point clouds in column 1. 
The adversarial point clouds are coloured according to the angle of displacement, $\boldsymbol{X}_{\text{adv}} - \boldsymbol{X}$. 
The heading pose error and the norm of the lateral and longitudinal pose errors of \ac{ICP} induced by each point cloud are indicated at the upper and lower right corners, respectively. The lateral and longitudinal errors are unitless as the point clouds are normalized.}
\label{fig:shapenet_fig}
\end{figure*}

\subsection{ShapeNetCore Results}
\label{shapenet_result}
\subsubsection{Implementation Details}
\label{shapenet_implementation}
We train our models on approximately 8,900 ShapeNetCore samples in batches of 32. We empirically determined that the optimal values for hyperparameters $a,b,c $ are $ 2,4,8 $, respectively. For $\mathcal{L}_{\text{adv}}$, we set $\alpha = 1$ and $\beta = 10$. We use the AdamW \cite{loshchilov2019decoupled} optimizer with a StepLR scheduler that reduces the initial learning rate of $10^{-4}$ by 30\% every 7 epochs. Before training, we pretrain the generator for 50 epochs on the training dataset with $\alpha = 0$ and $\beta = 1$, essentially asking it to reconstruct the original scans. Without pretraining, the generator generates scans that are too different from the maps to extract meaningful gradients from \ac{ICP}. We limit the maximum number of \ac{ICP} iterations to 25 to expedite training. For testing, we set the maximum number of iterations to 150, which we verify to be more than enough for \ac{ICP} to converge with an error tolerance of $10^{-4}$ in most cases. Samples where \ac{ICP} does not converge are dropped. 
The \ac{ICP} algorithm includes a Cauchy robust filter \cite{review_ICP} with the Cauchy parameter set to 0.15 and a maximum distance filter \cite{Segal2009GeneralizedICP} with $d_{max} = 0.3$, following common practices.

\subsubsection{Case Study on Weights in the Adversarial Loss}
\label{case_study}
This section examines how adversarial loss $\mathcal{L}_{\text{adv}}$ (\ref{eq:adv_loss}) weights affect the perturbations our model learns. We train our model on the ShapeNetCore dataset in three settings. In the $\theta-only$ setting, we set $w_6 = 1$ and all other weights to zero. The $x-only$ setting uses only the $x$-axis translation error in $\mathcal{L}_{\text{adv}}$ by setting $w_1 = 1$ and all other weights to 0. Finally, the $x-y$ setting uses both $x$-axis and $y$-axis translation errors in $\mathcal{L}_{\text{adv}}$ (i.e., $w_1 = w_2 = 1$ and  $w_3 = w_4 = w_5 = w_6 = 0$).
Adversarial point clouds produced by models trained in these three settings and the baselines are shown in Figure \ref{fig:shapenet_fig}. 
As expected, our model trained under the \textit{$\theta-only$} setting mainly causes rotations, inducing much higher rotational pose errors than other settings. The $x-only$ model, instead of rotating, shifts the original point clouds leftward, inducing $x$-axis pose errors. When we introduce $y$-axis pose errors into the adversarial loss in the $x-y$ setting, our model shifts the point clouds upward as well, in addition to leftward.

\setlength{\arrayrulewidth}{0.35pt}
\renewcommand{\arraystretch}{1.5} 

\begin{table*}[t]
  \fontsize{7}{8.7}\selectfont
  \centering
  \vskip 4pt
    \begin{tabularx}{\linewidth}{!{\vrule width 4\arrayrulewidth}*{1}{>{\centering\arraybackslash}m{.9cm}!{\vrule width 4\arrayrulewidth}}>{\centering\arraybackslash}m{1.4cm}*{4}{>{\centering\arraybackslash}m{1cm}|>{\centering\arraybackslash}m{1.4cm}}>{\centering\arraybackslash}m{1.04cm}!{\vrule width 4\arrayrulewidth}}
    \clineB{2-11}{4}
    \multicolumn{1}{c!{\vrule width 1.25pt}}{} & \multicolumn{2}{c|}{$\lambda = \SI{1}{\m}$} & \multicolumn{2}{c|}{$\lambda = \SI{2}{\m}$} & \multicolumn{2}{c|}{$\lambda = \SI{3}{\m}$} & \multicolumn{2}{c|}{$\lambda = \SI{4}{\m}$} & \multicolumn{2}{c!{\vrule width 1.25pt}}{$\lambda = \SI{5}{\m}$} \\
    \clineB{1-1}{4} \cline{2-11}
    \centering
    Method & Trans Pose Error [m] & \% Ours is Larger & Trans Pose Error [m] & \% Ours is Larger & Trans Pose Error [m] & \% Ours is Larger & Trans Pose Error [m] & \% Ours is Larger & Trans Pose Error [m] & \% Ours is Larger \\
    \Xhline{4\arrayrulewidth}
    Original & $0.07 \pm 0.21$ & 99.72\% &$0.07 \pm 0.21$ & 99.85\% &$0.07 \pm 0.21$ & 99.91\% &$0.07 \pm 0.21$& 99.94\% & $0.07 \pm 0.21$ & 99.96\%\\
    \hline
    Uniform & $1.16 \pm 0.16$ & 99.13\% & $2.17 \pm 0.14$ & 98.98\% & $3.14 \pm 0.18$ & 98.48\% & $4.12 \pm 0.29$ & 97.43\% & $5.07 \pm 0.48$ & 96.98\%\\
    \hline
    Normal & $1.31 \pm 0.20$ &98.17\% & $2.50 \pm 0.32$ & 96.65\% & $3.57 \pm 0.56$ & 94.58\% & $4.49 \pm 0.91$ & 90.07\% & $5.21 \pm 1.29$ & 88.28\%\\
    \hline
    Ours & \textbf{1.86 $\pm$ 0.48} & \text{-} & \textbf{3.21 $\pm$ 0.63} & \text{-} & \textbf{4.47 $\pm$ 0.79} & \text{-} & \textbf{5.49 $\pm$ 0.89} & \text{-} & \textbf{6.48 $\pm$ 1.00} & \text{-}\\
    \Xhline{4\arrayrulewidth}
  \end{tabularx}
  \caption{{Translation pose errors (higher is better) induced by our method and baselines under different perturbation bounds.} The ``\% Ours is Larger'' column lists the percentage of time our model induces a larger translation pose error than the models in comparison. Pose errors are listed in mean $\pm$ standard deviation.}
  \label{Table1}
\end{table*}
\subsection{Boreas Results}
\label{boreas_results}
\subsubsection{Implementation Details}
\label{boreas_implementaiton}
We train our models using 20,000 samples drawn from two Boreas repeat sequences associated with the same teach sequence. The training batch size is 6, with input point clouds to $\boldsymbol{G}$ sized $10,000 \times 3$. The optimal values of $a,b,c$ are found to be $3,12,48$, respectively. 
We set $\alpha = 1$ and $\beta = 10$ where $\alpha$ and $\beta$ are the weights of the adversarial and reconstruction losses, respectively, in the loss function. To be comparable to the baselines and \cite{laconte2023certifying}, we set $w_1 = w_2 = 1$ and $w_3 = w_4 = w_5 = w_6 = 0$ in the adversarial loss $\mathcal{L}_{\text{adv}}$. Consequently, for all the results in Section \ref{boreas_results}, only longitudinal and lateral pose errors are considered and reported. We train for 8 to 10 epochs ($\sim34$ hours) on a NVIDIA Tesla V100 32GB GPU, with an AdamW optimizer. 
During testing, it takes, on average, $\SI{1.67}{\s}$ to process one scan on the same GPU. However, as our method is offline, the training and inference time do not affect its usability. The learning rate is set to $10^{-4}$ initially and reduced by 30\% every 5 epochs. We also pre-trained $\boldsymbol{G}$ for 3 epochs.

The maximum number of \ac{ICP} iterations is capped at 25 to expedite training. For testing, we set the maximum number of iterations to 100, which is sufficient for \ac{ICP} to converge in most cases with a tolerance of $10^{-4}$. Test samples where \ac{ICP} does not converge are excluded. 
The \ac{ICP} algorithm includes a Cauchy filter (Cauchy parameter set to \SI{1}{\m}) and a maximum match distance filter ($d_{max} = \SI{5}{\m}$), following common practices. 

\subsubsection{Comparison with Baselines}
We test our attack and the baselines on four Boreas repeat sequences, which totals approximately $\SI{32}{\km}$ of driving data. We repeat this across five perturbation bounds and document the results in Table \ref{Table1}. Pose errors from \ac{ICP} localizing original scans (i.e., scans without adversarial perturbations) are also included for reference.
It is evident that our model induces significant pose errors in \ac{ICP} through adversarial perturbations. Moreover, our method consistently outperforms the baselines by a big margin across different perturbation bounds. Allowed the same amount of perturbation as the baselines, our method learns non-trivial perturbations that lead to higher pose errors at least 88\% of the time. This shows the efficacy of our method as an attack and a tool for worst-case analysis. 

\begin{figure}
    \centering
    \includegraphics[width=\linewidth]{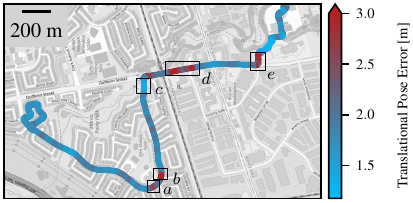}
    \caption{Worst-case translation pose errors over a route estimated by our model when allowed up to $\SI{1}{m}$ of perturbation. Locations are coloured based on their worst-case translation pose error. Errors are capped at \SI{3}{m} for better visualization.}
    \label{fig:worst-error-over-traj}
\end{figure}

\subsubsection{Pinpoint Dangerous Locations}
\label{Attack_trajectory}
This section demonstrates how our approach can use offline autonomous driving datasets to identify dangerous locations that autonomous robots should avoid when deployed. Locations are dangerous if perturbations in the measurements can lead to large \ac{ICP} errors.
We apply our model to corrupt live scans collected over three Boreas repeat sequences that follow the same route but under different weather conditions.
We plot the average translation errors induced at each location in Figure \ref{fig:worst-error-over-traj}. We set $\lambda = \SI{1}{\m}$ as perturbations of 1 meter are realistic and can lead to severe pose errors.
Locations in red are significantly more prone to attacks than others. Prior work \cite{laconte2023certifying} pinpoints very similar dangerous spots in this trajectory and shows that these spots correspond to locations particularly vulnerable to corrupted measurements.

Figure \ref{fig:worst-error-over-traj} identifies locations $a, b, d$, and $e$, all in the vicinity of open areas with few landmarks for localization, as dangerous. In contrast, location $c$, a tight suburban road with numerous houses and trees, is resilient to measurement corruptions.
For brevity, Figure \ref{fig:worst-error-over-traj} plots the average of different sequences. However, locations $a, b, d$, and $e$ are consistently
highlighted as dangerous, and location c is consistently labelled as safe, across sequences with different weather and road conditions. This indicates the generalizability of our offline findings for future deployments along the same route.

\subsubsection{Comparison with State of the Art}
\label{section:sota-comparison}
\begin{figure}[H]
  \centering
  \includegraphics[width=\linewidth]{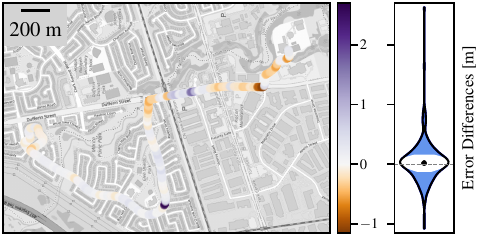}
  \caption{\ac{ICP} errors induced by our method minus those estimated by \cite{laconte2023certifying} on the same trajectory. A positive difference means ours finds a more detrimental perturbation pattern than \cite{laconte2023certifying}. The perturbation bound $\lambda = \SI{1}{m}$ for both approaches. Left: the trajectory coloured by the error difference. Right: a violin plot of the distribution of differences, with the interquartile range denoted in white.}
  \label{fig:sota}
\end{figure}

This section compares our method with \cite{laconte2023certifying} on a Boreas sequence. 
We extend \cite{laconte2023certifying} to perturb the entire point cloud to enable comparison. At each location in the sequence, we compare \ac{ICP} errors obtained from both methods, selecting the higher of the latitudinal and longitudinal errors, and visualize differences in the errors (ours minus \cite{laconte2023certifying}) in Figure \ref{fig:sota}. The violin plot shows that the two methods output similar errors in many places. Where they differ, \cite{laconte2023certifying} excels in structured environments (e.g., highways with adjacent buildings), whereas ours performs better in unstructured environments. Our method also outperforms \cite{laconte2023certifying} by a large margin at certain locations.

We theorize that, since \cite{laconte2023certifying} assumes single-iteration \ac{ICP} with known data associations, it fails to capture the complexity of \ac{ICP} in unstructured environments, as stated in their paper. Our method, which does not require these assumptions, can pinpoint dangerous unstructured environments that are prone to extreme, unusually high \ac{ICP} errors. As such, our method provides a valuable complement by detecting the particularly hazardous locations the existing method \cite{laconte2023certifying} could overlook.

\vspace{4pt}
\section{Conclusion}
    \label{sec:conclusion}
    In this paper, we propose the first learning-based attack against the widely used lidar-based \ac{ICP} algorithm as a tool for evaluating its resilience. Our attack learns to perturb a point cloud to maximize the pose error when using \ac{ICP} to localize it against a map. The attack induces significant \ac{ICP} errors and consistently outperforms baselines in more than 88\% of cases.
We demonstrate using the attack to estimate the worst pose errors \ac{ICP} may encounter during deployment, taking into consideration potential corruptions in the lidar measurements. In doing so, our approach can identify, pre-deployment, locations in a map with significant worst-case pose errors, highlighting vulnerabilities of \ac{ICP}. Autonomous robots can then avoid these risky locations, leading to safer deployments.
There is currently no enforcement that our model must generate perturbations that resemble any specific real-world events. For future work, we will investigate adding a discriminator to better simulate perturbations caused by real-world events.

\section*{ACKNOWLEDGMENT}
This work was supported by the OGS scholarship and ORF-RE program provided by the Province of Ontario.

\renewcommand*{\bibfont}{\footnotesize}
\printbibliography

\newpage
\appendix
\section{Appendix}
    \label{sec:appendix}
    \begin{figure*}[h]
    \includegraphics[width=\linewidth]{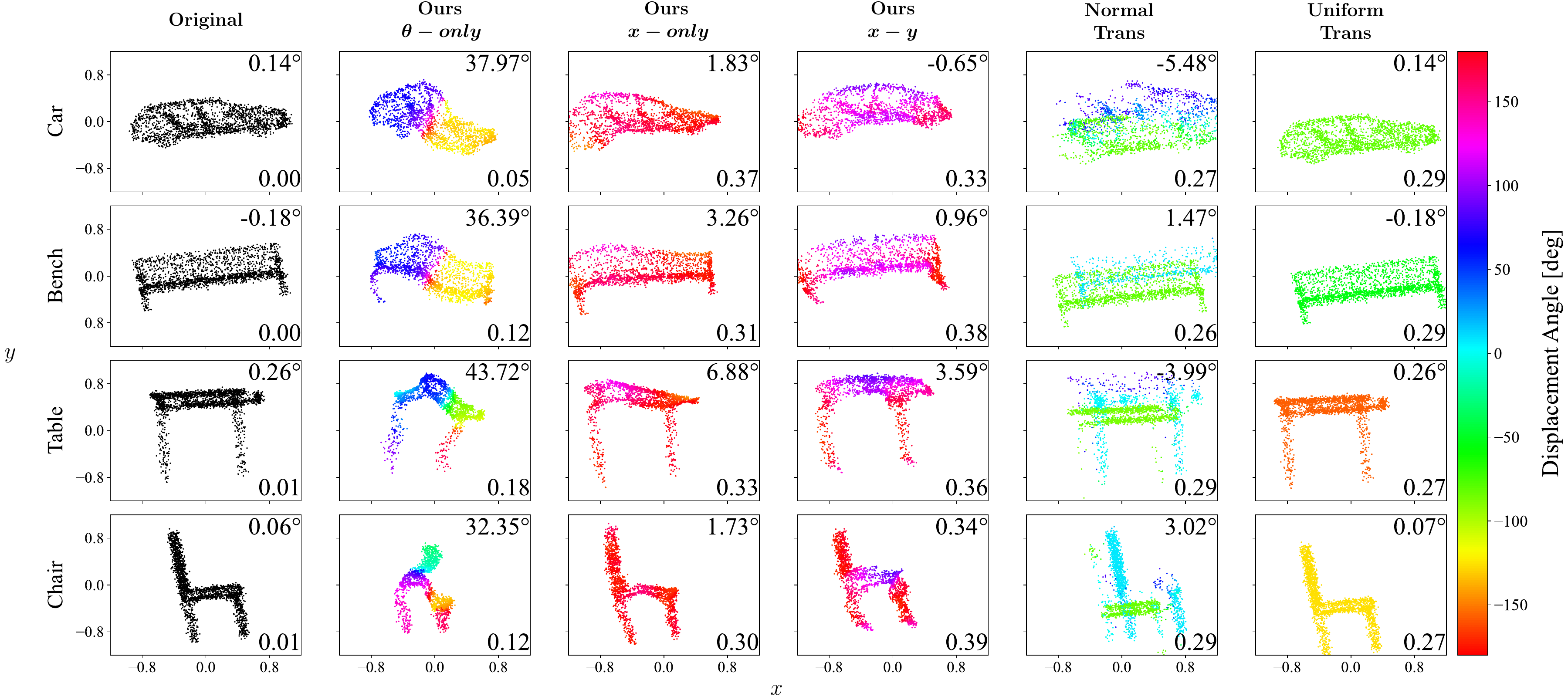}
\caption{An extended comparison between the original point clouds and the adversarial point clouds produced by our model and the baselines. Columns 2 through 6 showcase the adversarial point clouds resulting from applying our models and the baselines to the original point clouds in column 1. 
The adversarial point clouds are coloured according to the angle of the displacement vectors, $\boldsymbol{X}_{\text{adv}} - \boldsymbol{X}$. 
The heading pose error and the norm of the lateral and longitudinal pose errors of \ac{ICP} induced by each point cloud are indicated at the upper and lower right corners, respectively. The lateral and longitudinal pose errors are unitless as the point clouds are normalized. Minor \ac{ICP} pose errors are observed even when localizing the uncorrupted point clouds in column 1. This occurs because noise is added to the scans to prevent perfect alignment between the scans and the maps.}
\label{fig:shapenet_fig_extended}
\end{figure*}

\begin{figure}[h]
    \includegraphics[width=\linewidth]{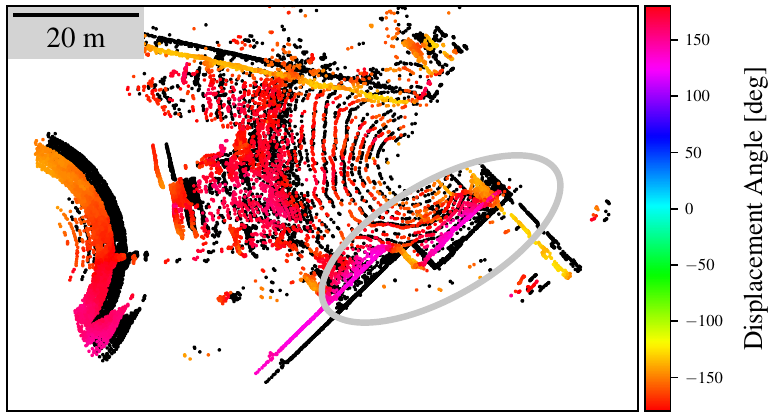}
\caption{An example of the adversarial perturbations our model learns on the Boreas dataset \cite{burnett_ijrr23}. The adversarial point cloud generated by our model (coloured according to the angle of the displacement vector) is overlaid on top of the original point
cloud (in black) for contrast. The linear features circled in grey are shifted in such a way that the features stay connected.}
\label{fig:example_boreas_result}
\end{figure}

Extended ShapeNetCore \cite{chang2015shapenet} results are shown in Figure \ref{fig:shapenet_fig_extended}. Additionally, Figure \ref{fig:example_boreas_result} shows an example of the adversarial point clouds generated by our model.
\end{document}